\documentclass{article}

    \PassOptionsToPackage{round, compress}{natbib}
\usepackage[preprint]{neurips_2024}
\usepackage[utf8]{inputenc}              % allow utf-8 input
\usepackage[T1]{fontenc}                 % use 8-bit T1 fonts
\usepackage{hyperref}                    % hyperlinks
\usepackage{url}                         % simple URL typesetting
\usepackage{booktabs}                    % professional-quality tables
\usepackage{amsfonts}                    % blackboard math symbols
\usepackage{nicefrac}                    % compact symbols for 1/2, etc.
\usepackage{microtype}                   % microtypography
\usepackage{xcolor}                      % colors
\usepackage{multirow}                    % enable merge table cells
\usepackage{graphicx}                    % enhance support for graphs
\usepackage[inkscapelatex=false]{svg}    % enable svg image type
\usepackage{amsmath}                     % to use \eqref

\usepackage{amsmath}
\usepackage{amsfonts}
\usepackage{amssymb}
\usepackage{wrapfig}
\usepackage{subcaption}
\usepackage{multirow}
 \usepackage{mathtools} 

\usepackage{verbatim}

\usepackage{anyfontsize}

\usepackage{microtype}
\usepackage{graphicx}
\usepackage{booktabs} % for professional tables
\usepackage{multirow}
\usepackage{amsmath,amssymb}
\usepackage{booktabs}
\usepackage{caption,subcaption}

\usepackage{xcolor}
\definecolor{mygreen}{HTML}{3cb44b}
\definecolor{skyblue}{HTML}{beffff}
\definecolor{lightgreen}{HTML}{90ee90}

\usepackage{color, colortbl}

\definecolor{emerald}{rgb}{0.31, 0.78, 0.37}

\usepackage{tcolorbox}
\usepackage{enumitem}
\setitemize{itemsep=10pt,topsep=0pt,parsep=0pt,partopsep=0pt}
\usepackage{colortbl}

\usepackage{xcolor}
\definecolor{mygreen}{HTML}{3cb44b}
\colorlet{myyellow}{green!10!orange!90!}
\makeatletter

\usepackage{tikz}
\usetikzlibrary{arrows,shapes,automata,backgrounds,fit,petri}
\usepackage{adjustbox}

\newcommand{\RN}[1]{%
	\textup{\lowercase\expandafter{\it \romannumeral#1}}%
}
\usepackage{tabu}
\newcommand{\eg}[0]{\emph{e.g., }}

\newcommand{\beq}{\vspace{0mm}\begin{equation}}
\newcommand{\eeq}{\vspace{0mm}\end{equation}}
\newcommand{\beqs}{\vspace{0mm}\begin{eqnarray}}
\newcommand{\eeqs}{\vspace{0mm}\end{eqnarray}}
\newcommand{\barr}{\begin{array}}
\newcommand{\earr}{\end{array}}

\usepackage{color, colortbl}
\definecolor{Gray}{gray}{0.93}

\usepackage{lipsum}

\usepackage{pifont}% http://ctan.org/pkg/pifont

\usepackage{makecell}

\usepackage{xcolor,amsmath}
\usepackage[linesnumbered,ruled,vlined]{algorithm2e}
\DontPrintSemicolon

\usepackage{xcolor}
\definecolor{mygreen}{HTML}{3cb44b}

\SetKwComment{Comment}{\color{green!50!black}\# }{}

\SetKwProg{Function}{def}{:}{}

\SetKwProg{For}{for}{:}{}
\SetKwProg{If}{if}{:}{}

\newcommand{\tablestyle}[2]{\setlength{\tabcolsep}{#1}\renewcommand{\arraystretch}{#2}\centering\footnotesize}

\hypersetup{
  colorlinks   = true, %Colours links instead of ugly boxes
  urlcolor     = blue, %Colour for external hyperlinks
  linkcolor    = blue, %Colour of internal links
  citecolor   = violet %Colour of citations
}

\newcommand{\mypm}{\mathbin{\mathpalette\@mypm\relax}}
\newcommand{\@mypm}[2]{\ooalign{%
  \raisebox{.1\height}{$#1+$}\cr
  \smash{\raisebox{-.6\height}{$#1-$}}\cr}}
  
\newif\ifshowcomments
\showcommentstrue
\ifshowcomments
\newcommand{\mynote}[2]{\textcolor{blue}{\fbox{\bfseries\sffamily\scriptsize#1}}
  \textcolor{blue}{{$/*$\textsf{\emph{#2}}$*/$}}}
\ifshowcomments

\else
\newcommand{\mynote}[2]{}
\fi
\newcommand{\llavaname}{LLaVA-Med}
\newcommand{\shortname}{Llama3-Med}
\newcommand{\eat}[1]{\ignorespaces}

\title{Advancing High Resolution Vision-Language Models in Biomedicine}

\author{%
  Zekai~Chen~~~~Arda~Pekis~~~~Kevin~Brown \\
  Standard Model Biomedicine, Inc. \\
  \texttt{\{zach,arda,kevin\}@standardmodel.bio} \\
}

\begin{document}

\maketitle

\begin{abstract}
  % Multi-modal learning has revolutionized generative AI, particularly in vision-language modeling. Advances like multi-modal GPT-4V and open source projects such as LLaVA have enabled robust conversational agents capable of zero-shot task completion. However, applying these technologies to the biomedical field presents unique challenges. Recent efforts, like LLaVA-Med, have extended instruction-tuning to biomedical domains using large-scale datasets like PMC-15M. Our work introduces three key contributions: (i) a new instruct dataset enriched with medical image-text pairs from Claude3-Opus and LLaMA3 70B, (ii) an innovative image encoding strategy using hierarchical representations for fine-grained biomedical visual understanding, and (iii) the Llama3-Med model, which achieved state-of-the-art performance on biomedical visual question answering benchmarks with an improvement over 10\% on average compared to previous methods. These advancements offer more accurate, reliable tools for medical professionals, addressing gaps in current multi-modal conversational assistants and paving the way for further medical AI innovations.

Multi-modal learning has transformed generative AI, particularly in vision-language modeling. Advances such as the multi-modal GPT-4V and open-source projects like LLaVA have enabled robust conversational agents capable of zero-shot task completions. However, extending these technologies in the biomedical field introduces unique challenges. Recent initiatives like LLaVA-Med have begun to tailor instruction-tuning to biomedical contexts using extensive datasets like PMC-15M. Our research contributes three significant advancements: (i) we introduce a new instruct dataset enriched with medical image-text pairs derived from Claude3-Opus and LLaMA3 70B, (ii) we propose an innovative image encoding strategy that employs hierarchical representations to enhance fine-grained biomedical visual comprehension, and (iii) we develop the Llama3-Med model, which achieves state-of-the-art zero-shot performance on biomedical visual question answering benchmarks, improving performance by over 10\% on average compared to prior methods. These advancements provide more precise and reliable tools for medical professionals, effectively bridging gaps in current multi-modal conversational assistants and fostering further innovations in medical AI. Codes available at \url{https://github.com/standardmodelbio/Llama3-Med.git}.

\end{abstract}

\section{Introduction}
\label{sec:intro}

The integration of multimodal data through auto-regression has emerged as a transformative approach in artificial intelligence, driving advances in general-domain multimodal modeling~\citep{tsimpoukelli2021multimodal,driess2023palm}. Pioneering works leveraging generative pretraining techniques, exemplified by multi-modal GPT-4V~\citep{gpt4} and open projects such as LLaVA~\citep{liu2023visual} and Qwen~\citep{Bai2023QwenTR}, have demonstrated significant capabilities in handling complex, multimodal inputs. These advancements facilitate the development of conversational agents capable of robust zero-shot task completion across diverse vision-language tasks~\citep{Zhang2023VisionLanguageMF}, highlighting the potential for multimodal models to serve as general-purpose conversational assistants. The seamless integration of visual and textual data empowers these models to understand and generate coherent, contextually relevant responses, advancing AI capabilities across various applications.

% Despite these successes, the application of such technologies to the biomedical field has been challenging. The unique characteristics of biomedical image-text pairs demand a more specialized approach than what general-domain models can offer. Recent efforts have extended the utility of instruction-tuning to the biomedical domain, leveraging large-scale scientific datasets like PMC-15M~\citep{zhang2023large} to train models capable of more sophisticated biomedical visual understanding. LLaVA-Med~\citep{Li2023LLaVAMedTA}, as a representative, presented a pioneering effort to extend multi-modal instruction-tuning to the biomedical domain, leveraging GPT-4 to generate diverse biomedical multi-modal instruction-following data using image-text pairs from the large-scale PMC-15M dataset~\citep{zhang2023large} and fine-tuned a biomedical vision-language model using a curriculum learning method~\citep{liu2023visual}. These innovations have highlighted the potential of tailored multi-modal models to address the nuanced requirements of biomedical data.

While these advancements are promising, applying such technologies to the biomedical field presents distinct challenges. The unique characteristics of biomedical image-text pairs demand more specialized models than those developed for general domains. Recent initiatives like LLaVA-Med~\citep{Li2023LLaVAMedTA} have specifically adapted instruction-tuning for the biomedical context, utilizing extensive scientific datasets such as PMC-15M~\citep{zhang2023large} to enhance models' capabilities in sophisticated biomedical visual understanding. This approach involves using GPT-4 to generate a diverse array of biomedical multi-modal instruction-following data from PMC-15M~\citep{zhang2023large} image-text pairs, followed by fine-tuning a vision-language model through a curriculum learning strategy. Such efforts underscore the potential of customized multi-modal models to meet the intricate demands of biomedical applications, demonstrating significant progress in this specialized area.

Expanding upon the promising direction of LLaVA-Med, our work presents three key contributions in the field. First, we have curated a new instruct dataset enriched with medical image-text pairs, collaboratively generated by the cutting-edge Claude3-Opus~\citep{TheC3} and LLaMA3 70B~\citep{llama3} models instead of using GPT-4 solely.
This dataset significantly broadens the scope of samples beyond those in existing collections, such as \emph{llava-med-instruct-60k}~\citep{Li2023LLaVAMedTA}, providing a robust supplementary resource for model training.
By enhancing the diversity and richness of our training materials, this dataset incorporates a more varied and representative selection of biomedical imagery and textual data.
% This dataset covers a significantly broader range of samples compared to the existing data, building on the foundation provided by PMC-15M and serving as a powerful complementary resource for model training. 
% The expanded dataset not only enhances the diversity and richness of the training material but also addresses the limitations of existing datasets by incorporating more varied and representative samples of biomedical imagery and text.

Second, drawing inspiration from models such as MM1~\citep{McKinzie2024MM1MA}, LLaVA-Next~\citep{liu2024llavanext} and Mini-Gemini~\citep{Li2024MiniGeminiMT}, we implement an innovative image encoding strategy that employs hierarchical representations across various resolutions, from high to low.
This method enhances the granularity with which medical images are analyzed, a crucial improvement given the intricate and detailed nature of biomedical imagery. \emph{By capturing subtle nuances across multiple scales, our models gain the ability to interpret visual data with greater depth and precision}. This hierarchical encoding strategy ensures that vital details are preserved, thereby enabling the models to deliver more precise and contextually appropriate responses in biomedical settings.
% for fine-grained biomedical visual understanding. This approach allows for a more granular understanding of medical images, which is critical given the complexity and detailed nature of biomedical imagery. By capturing nuances at multiple scales, our models can achieve a deeper and more accurate interpretation of visual data. The hierarchical encoding strategy ensures that important details are not lost, enabling the model to provide more precise and contextually relevant responses in biomedical applications.

Finally, we introduce the Llama3-Med model, which leverages our enriched dataset and advanced encoding techniques, built upon the robust foundation provided by the acclaimed open-source large language model (LLM), LLaMA3~\citep{llama3}.
Llama3-Med has achieved unprecedented state-of-the-art (SoTA) performance on well-established visual question answering benchmarks, such as VQA-RAD~\citep{lau2018dataset}, VQA-PATH~\citep{he2020pathvqa} and SLAKE~\citep{Liu2021SlakeAS}, in the biomedical domain. This notable achievement not only demonstrates the effectiveness of our methodological innovations but also establishes a new benchmark for future research in this field. The exemplary performance of Llama3-Med underscores its potential to transform the application of AI in the biomedical sector, offering tools that are more precise, reliable, and adept at navigating complex medical scenarios.
\section{Better Perception of Biomedical Images with Higher Resolutions}
\label{sec:method}

% \paragraph{Hierarchical Representation Learning of Biomedical Images.}
\begin{figure}[t]
% \vspace{-4mm}
\centering
\includegraphics[width=\linewidth]{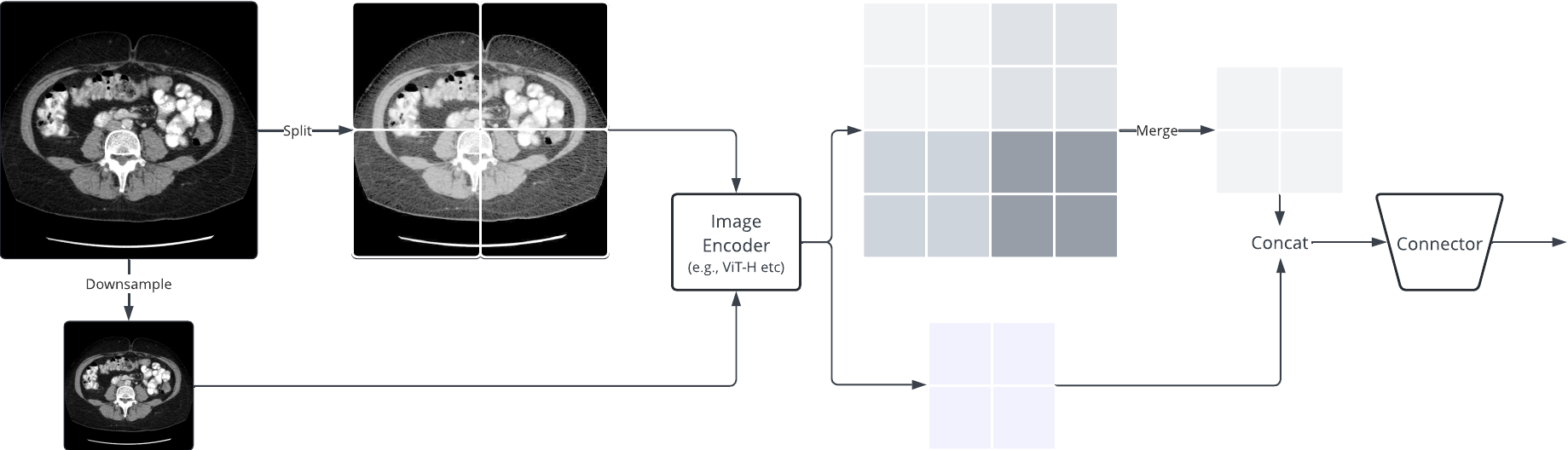}
% \vspace{1mm}
\caption{Illustration of building feature embedding in \shortname{}. High resolution biomedical images are split into multiple smaller pieces that are digestible by existing CLIP image encoders~\citep{radford2021learning}. Embeddings of hierarchical representations are further concatenated and fed into connector for fine-tuning.}
\label{fig:s2}  
\vspace{-5mm}
\end{figure}

Image resolution plays a pivotal role in biomedical image encoding and vision-language understanding, as it directly influences the ability to detect fine structures and subtle abnormalities in medical images~\citep{Chen_2023_WACV}. High-resolution images are essential for multi-scale analysis and are particularly effective when combined with advanced imaging technologies like MRI and CT scans, ensuring consistency in quality across different scales. This high level of detail is crucial for vision-language models, as improved image resolution enhances the quality of visual input. This, in turn, boosts the performance of language-based tasks, such as automatic report generation and interactive medical systems, leading to more accurate and reliable outcomes in medical applications.

Inheriting the innovative spirit of prior works~\citep{McKinzie2024MM1MA,liu2024llavanext,Shi2024WhenDW}, we adopt a hierarchical representation learning approach that optimizes the performance of existing well pre-trained vision encoders~\citep{zhang2023large}. 
Rather than increasing model size, we enhance fine-grained signal perception by processing higher resolution images across multiple scales.
\begin{wrapfigure}{r}{0.45\textwidth}
% \vspace{-4mm}
\centering
\includegraphics[width=\linewidth]{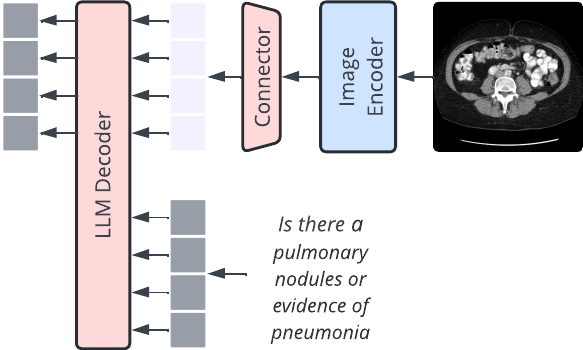}
% \vspace{1mm}
\caption{Illustration of instruction fine-tuning paradigm. Similar to~\citet{Li2023LLaVAMedTA}, we freeze the image encoder while fine-tuning the connector and LLM base.}
\label{fig:finetuning}  
  \vspace{-5mm}
\end{wrapfigure}
Specifically, this method allows a pre-trained vision encoder (\eg CLIP~\citep{radford2021learning} or SigLIP~\citep{Zhai2023SigmoidLF}) to encode images at different scales (\eg $756^2$ and $1134^2$) by segmenting larger scale images into smaller sub-images.
These are then processed and reassembled to ensure feature consistency across scales, as illustrated in Figure~\ref{fig:s2}. 
This technique maintains the model’s original complexity while significantly improving its feature extraction capabilities. By avoiding the computational burden of larger vision encoders and eliminating the need for additional pre-training, this design achieves both efficiency and scalability.

While most existing CLIP encoders support resolutions ranging from 224 to 336, these are often inadequate for fully capturing the intricacies of biomedical images. Recognizing the critical role of resolution in this context, we have adopted the use of the encoder pre-trained with a CLIP objective on DFN-5B~\citep{Fang2023DataFN}, which supports a higher resolution of 378×378\footnote{https://huggingface.co/apple/DFN5B-CLIP-ViT-H-14}. This enhancement allows for more detailed and accurate modeling of biomedical imagery.

\vspace{-2mm}
\paragraph{Overall Training Paradigm.}
In line with the established vision-language modeling framework~\citep{liu2023visual,Li2023LLaVAMedTA,McKinzie2024MM1MA} that combines an image encoder with a vision-language connector to integrate visual and textual data, we have implemented a consistent overall training paradigm. This adaptation employs a structured two-stage training process to ensure thorough integration of biomedical image and text data: (i) \emph{Vision-Language Connector Pre-training Stage.} During this initial stage, image-text pairs are transformed into instruction-following data. The model tasks involve describing the image and attempting to predict its original caption, with the visual encoder and LLM weights remaining frozen, and only the connector layers being updated. This phase is crucial for aligning the image features with textual word embeddings from the pre-trained LLM, thereby broadening the biomedical vocabulary and enhancing the stability for subsequent fine-tuning. (ii) \emph{End-to-End Instruction Fine-tuning.} Illustrated in Figure~\ref{fig:finetuning}, this stage involves unfreezing the LLM and connector weights while maintaining the visual encoder as frozen. The focus is on fine-tuning the model using biomedical language-image instruction-following data. This stage significantly boosts the model's capabilities as a biomedical visual assistant and enhances its performance on zero-shot tasks within established biomedical Visual Question Answering (VQA) datasets. This dual-phase approach meticulously prepares the model to handle complex biomedical queries, optimizing its effectiveness and responsiveness in real-world applications.

Further supervised fine-tuning can be applied to specific biomedical scenarios using various Biomed-VQA datasets, allowing for the development of precise, tailored models. This approach is not only cost-effective, as evidenced by reasonable development and operational expenses, especially considering the training duration on specialized hardware, but it also highlights the model’s versatility across different domains. Such scalability promotes continuous instruction-tuning for domain-specific applications. Moreover, the network's flexibility facilitates potential performance improvements by integrating components from domain-specific models, such as various CLIP~\citep{radford2021learning} image encoders~\citep{zhang2023large}. This adaptable training paradigm has been chosen due to its cost efficiency and proven success, consistently achieving high performance in related tasks.

% \vspace{-3mm}

\paragraph{Rationale of Data Synthesis.} 
LLaVA-Med~\citep{Li2023LLaVAMedTA} developed a groundbreaking dataset from PMC-15M~\citep{zhang2023large} through a combined machine-human curation effort. This dataset comprises two main components: (i) \emph{The Concept Alignment Data}, featuring 600K image-text pairs from PMC-15M~\citep{zhang2023large}, used to generate single-round instruction-following examples by posing questions related to the images and their captions. (ii) \emph{The Instruction-Tuning Data} includes multi-round conversations generated by GPT-4, rephrasing context from PubMed papers to conversational format, spanning 60K image-text pairs across various imaging modalities such as CXR, CT, MRI, histopathology, and gross pathology.

\begin{wraptable}{l}{8cm}
% \centering 
\vspace{-4mm}
\tablestyle{2.5pt}{1.2}
\begin{tabular}{@{}lccccc@{}}  
\toprule
  & Chest X-Ray & CT & MRI & Histology & Pathology \\
  \midrule
  \multicolumn{6}{l}{\it \llavaname{} (100\% GPT4)} \\
  \# Images & 8,060 & 17,358 & 16,539 & 17,030 & 2,552 \\
  \# QAs & 22,839 & 50,583 & 47,867 & 49,504 & 7,323 \\ \midrule
  \multicolumn{6}{l}{$\Delta$ \it \shortname{} (25\% Claude3-Opus \& 75\% LLaMA3 70B)} \\
  \rowcolor{Gray}
  \# Images & 40 & 6,809 & 6,568 & 6,643 & 116 \\
  \rowcolor{Gray}
  \# QAs & 212 & 41,630 & 40,824 & 42,084 & 714 \\
 \bottomrule 
\end{tabular}
% \vspace{1mm}
\caption{The statistical overview of Llama3-Med instruction data compared to LLaVA-Med: distribution of number of images and QA turns across five domains.}
\label{tab:finetuning_data_stats}  
\vspace{-2mm}
\end{wraptable}
In the process of concept alignment, we discovered that merely adding more data does not significantly enhance fine-tuning performance, as the static nature of the image encoder and LLM base leads to quick saturation of the connector. Consequently, to improve this dataset, we incorporated synthetic data generated by the latest SoTA Claude3-Opus~\citep{TheC3} and LLaMA3 70B~\citep{llama3} models, specifically for instruction fine-tuning. 
Table~\ref{tab:finetuning_data_stats} provides a statistical comparison of our enhanced dataset against LLaVA-Med. Our strategic use of a 25\% Claude3 and 75\% LLaMA3 70B mix resulted in fewer images across most domains but maintained a robust count of QA pairs, especially in CT, MRI, and Histology. This approach highlights our \emph{strategic emphasis on enriching the quality of QA pairs rather than merely expanding the volume of images}.

Utilizing both Claude3~\citep{TheC3} and LLaMA3~\citep{llama3}, instead of relying solely on GPT-4, provides significant advantages. \emph{LLaMA native generation aligns closely with our model architecture}, \emph{ensuring coherence and consistency} through \emph{soft knowledge distillation} which intangibly enhance fine-tuning efficiency. On the other hand, Claude3, originating from a different training background, introduces valuable diversity and helps mitigate potential biases~\citep{Zack2024AssessingTP} inherent in training with a single model. This dual-model strategy reduces the risk of bias and improves the robustness and versatility of our dataset, thereby leading to the development of more accurate and adaptable biomedical vision-language models. (\emph{w/ Details in Supplemental Materials})

\section{Experiments}
\label{sec:exp}

We conducted comprehensive experiments to assess \shortname{}'s performance on standard benchmarks against existing methods, as well as the quality of its open generation on multimodal biomedical instructions.

\begin{wraptable}{r}{8cm}
\vspace{-5mm}
\centering 
\tablestyle{2pt}{1.2}
% \centering 
% \tablestyle{4pt}{1.2}
\begin{tabular}{@{}l|cc|ccc|ccc@{}}  
\toprule
  & \multicolumn{2}{c|}{\bf VQA-RAD} & \multicolumn{3}{c|}{\bf SLAKE} & \multicolumn{3}{c}{\bf PathVQA}  \\
Dataset   & Train    & Test   & Train  & Val  & Test    & Train  & Val & Test  \\ \midrule
\# Images & 313  &  203 &   450  & 96 &  96 & 2599 & 858 & 858 \\
 \# QA Pairs & 1797   & 451   & 4919  &  1053    &  1061 & 19,755   & 6279   & 6761   \\
 \# Open &   770  &  179  &   2976  &  631   &    645   &   9949  &   3144 &  3370 \\
 \# Closed &   1027 &  272 &     1943 &  422  &     416  &    9806 &    3135 &  3391 \\
 \bottomrule 
\end{tabular}
% \vspace{1mm}
\caption{Dataset statistics. For SLAKE, we focus exclusively on the English subset when making direct comparisons with existing methods.}
\label{tab:data_stat_three_datasets}  
\vspace{-5mm}
\end{wraptable}

% \vspace{-4mm}
\paragraph{Datasets for Evaluation and Benchmarking.}
Three diverse biomedical VQA datasets were utilized for the evaluation of \shortname{}, as follows:  
\textbf{\textit{VQA-RAD}}~\citep{Lau2018DescriptorA} includes 3,515 QA pairs generated by clinicians across 315 radiology images, distributed evenly across the head, chest, and abdomen. Questions are categorized into 11 types (\eg abnormality, attribute, modality) and are split between closed-ended (58\%, yes/no or limited choices) and open-ended (42\%) formats.
\textbf{\textit{VQA-PATH}}~\citep{He2020PathVQA3Q} comprises 4,998 pathology images coupled with 32,799 QA pairs, addressing multiple aspects such as location, shape, color, and appearance. This dataset differentiates between open-ended (\eg why, what, how, where) and closed-ended questions.
\textbf{\textit{SLAKE}}~\citep{Liu2021SlakeAS} features 642 radiology images and over 7,000 QA pairs annotated by experienced physicians. This dataset is enriched with visual annotations such as semantic segmentation masks and object detection bounding boxes, covering a broad range of body parts. Although SLAKE is a bilingual dataset, only the English subset was used for our analysis.
All datasets were de-identified, ensuring compliance with privacy standards. Detailed statistics for these datasets are summarized in Table~\ref{tab:data_stat_three_datasets}, highlighting \shortname{}’s comprehensive approach to evaluation across varied medical VQA benchmarks.

\paragraph{Evaluation Metrics.}
For closed-set questions, we report \emph{accuracy} to assess correctness. For open-set questions, we use \emph{recall} to measure the proportion of ground-truth tokens that appear in the generated responses, ensuring a fair comparison with \llavaname{}. In the existing literature, responses found in the training set are often used as candidate answers for testing. However, our approach does not constrain the responses to open-set questions, adhering more closely to their inherently open-ended nature. This makes our method more challenging yet more reflective of real-world applications.

\paragraph{Baselines.}
\shortname{} is also compared with previous SoTAs such as VL Encoder–Decoder~\citep{bazi2023vision}, Q2ATransformer~\citep{liu2023q2atransformer}, Prefix T. Medical LM~\citep{van2023open}, PubMedCLIP~\citep{eslami2023pubmedclip}, BiomedCLIP~\citep{zhang2023large} and M2I2~\citep{li2022self}. Moreover, for fair comparison, we trained an independent LLaVA-Med native model from scratch, however, using Mistral-7B~\citep{Jiang2023Mistral7} as the foundation model instead of Vicuna~\citep{zheng2023judging}. We will present the performance of this model in the results section.

\paragraph{Implementation details.} All experiments were conducted on a high-performance setup featuring four NVIDIA H100 GPUs. During pre-training, we utilized a batch size of 256 tailored for maximum efficiency and an initial learning rate of $1e^{-3}$ with cosine scheduler, optimized via the AdamW. The fine-tuning phase was adjusted with a batch size of 128 and an initial learning rate of $2e^{-5}$. Our computational environment, built on PyTorch version 2.1 and LLaVA code base~\citep{liu2024llavanext}. For detailed configurations, code snippets, and additional computational optimizations, please refer to Appendix~\ref{sec:appendix}.

\begin{table}[t]
\centering
\vspace{-5mm}
\tablestyle{2pt}{1.1}
\begin{tabular}{ll|ccc|ccc|ccc|c}  
\toprule
 & & \multicolumn{3}{c|}{\bf VQA-RAD} & \multicolumn{3}{c|}{\bf SLAKE} & \multicolumn{3}{c|}{\bf VQA-PATH} & \multirow{2}{*}{Average} \\
Method   &  & Ref   & Open   & Closed   & Ref   & Open   & Closed    & Ref  & Open &  Closed &  \\ \midrule
 \multicolumn{11}{l}{\it Existing SoTA methods (reported by literature) } \\  
\multicolumn{2}{l|}{\citep{bazi2023vision}} &  71.49 & / & 82.47 & / & / & / & 71.49 & / & 85.61 & / \\
\multicolumn{2}{l|}{\citep{liu2023q2atransformer}} &  79.19 & / & 81.20 & / & / & / & 54.85 & / & 88.85 & / \\
\multicolumn{2}{l|}{\citep{van2023open}} & / & / & / & 84.30 & / & 82.01 & 40.00 & & 87.00 & / \\ 
\multicolumn{2}{l|}{\citep{eslami2023pubmedclip}} &  60.10 & / & 80.00 & 78.40 & / & 82.50 & / & / & / & / \\
\multicolumn{2}{l|}{\citep{zhang2023large}} &  67.60 & / & 79.80 & 82.05 & / & 89.70 & / & / & / & / \\
\multicolumn{2}{l|}{\citep{li2022self}} & 66.50 & / & 83.50 & 74.70 & / & 91.10 & 36.30 & / & 88.00 & / \\
\midrule
\multicolumn{11}{l}{\it Supervised fine-tuning results of 1 epoch} \\
   % \rowcolor{Gray}
\multicolumn{2}{l|}{\llavaname{} 7B (CLIP)} &  / & 28.61 & 56.25 & / & 70.58 & 54.57  & / &  11.17   & 59.19 & 46.73 \\ 
\multicolumn{11}{l}{\it Supervised fine-tuning results of 9 epochs} \\
\multicolumn{2}{l|}{LLaVA 7B} &  / & 50.00 & 65.07 & / & 78.18 & 63.22 & / & 7.74 & 63.20 & 54.57 \\ 
\multicolumn{2}{l|}{\llavaname{} 7B (CLIP)} &  / & 66.26 & 80.88 & / & 82.30 & 84.86  & / &  37.59   & 91.54 & 73.90 \\ 
   % \rowcolor{Gray}
\multicolumn{2}{l|}{\llavaname{} 7B (BioMed CLIP)} & / & 64.75 & 83.09 & / & 87.11 & 86.78 & / & 39.60 & 91.09 & 75.40 \\ 
 % \rowcolor{Gray}
\multicolumn{2}{l|}{\llavaname{} 13B (CLIP)} & / & 64.58 & 77.94 & / & 84.97 & 85.58  & / &  38.82  & 92.39 & 74.05 \\ 

\midrule

\multicolumn{11}{l}{\it Zero-shot results w/o supervised finetuning } \\
\multicolumn{2}{l|}{LLaVA 7B} &  / & 20.74 & 59.19 & / & 26.82 & 50.24 & / & 8.74 & 45.65 & 35.23 \\  
   % \rowcolor{Gray}
\multicolumn{2}{l|}{\llavaname{} 7B (CLIP)} &  / & 28.23 & 61.40 & / & 39.17 & 52.16 & / & 12.30 & 54.05 & 41.22 \\
   % \rowcolor{Gray}
\multicolumn{2}{l|}{\llavaname{} 7B (BioMed CLIP)} & / & 37.84 & 60.66 & / & 39.73 & 54.33 & / & 11.65 & 49.07 & 42.21 \\ 
 % \rowcolor{Gray}
\multicolumn{2}{l|}{\llavaname{} 13B (CLIP)} &  / & 31.66 & 61.40 & / & 37.71 & 49.76 & / & 11.34 & 49.63 & 40.25 \\ 
\multicolumn{11}{l}{\it Our runs } \\
    \rowcolor{Gray}
\multicolumn{2}{l|}{\llavaname{} 7B (w/ Mistral)} &  / & 32.03 & \underline{62.87} & / & \underline{41.63} & \underline{56.97} & / & 11.80 & \underline{55.47} & \underline{43.46} \\ 
    \rowcolor{Gray}
\multicolumn{2}{l|}{\shortname{} 8B} &  / & 31.20 & {\bf 62.87} & / & {\bf 44.34} & {\bf 61.06} & /& {\bf 12.88} & {\bf 64.97} & {\bf 46.22} \\

\bottomrule
\end{tabular}
\vspace{1mm}
\caption{Comparison with previous state-of-the-art supervised methods. For open-ended questions, previous approaches address the problem as a classification task among predefined answers in the training set. This strategy may overstate their generalizability, as these datasets usually include test answers that are frequently found in the training data.
}
\vspace{-5mm}
\label{tab:results_on_three_datasets}
\end{table} 

\subsection{Results on Biomedical VQAs}

\begin{table}[t]
  \begin{minipage}{0.99\textwidth}
\centering  
\small
\vspace{-5mm}
\scalebox{0.83}{
\begin{tabular}{p{2.1cm} p{13.2cm}  }
\toprule
\multicolumn{2}{l}{\bf Chest X-Ray example from~\citet{bharat2020vaping}}  \\
\midrule
% Caption & Chest x-ray (CXR) on Day 2 of admission post-intubation (yellow line showing the level of the endotracheal tube). Rapidly worsening ground-glass opacities are seen throughout the lungs with relative subpleural sparing (red arrows). \citep{bharat2020vaping}
% &  \includegraphics[height=3.5cm]{figures/example_imgs/visual_chat_example.jpg} \\
&  \includegraphics[height=4cm]{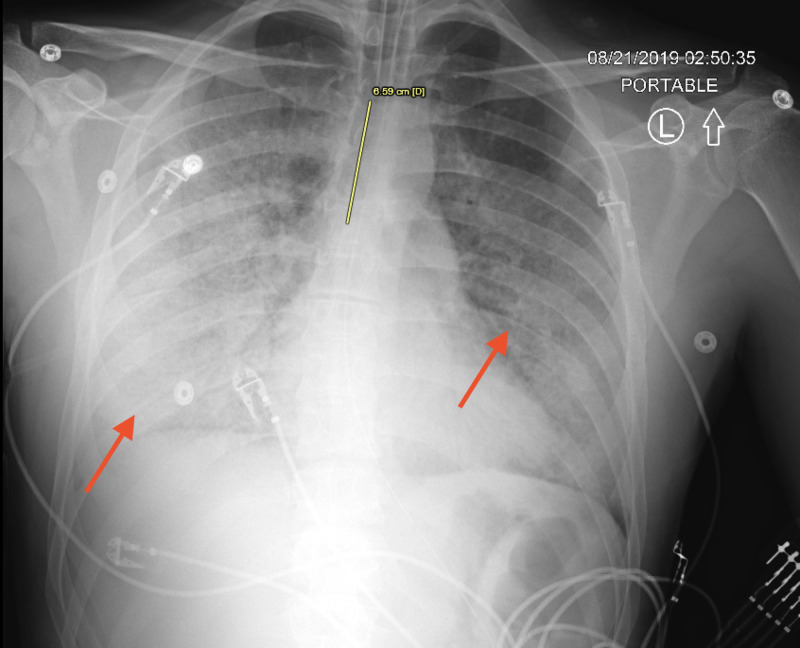} \\
 & {\tiny Source:~\url{https://www.ncbi.nlm.nih.gov/pmc/articles/PMC7398722/}} \\
 User & What is shown in this image?  \\
LLaVA &  The image shows a ribcage with wires coming out of it, indicating that it is likely a medical illustration or diagram.  \\
User &  Are there any devices implanted in this patient?  \\
LLaVA  &  Yes, there are several devices protruding from the patient's chest, which are likely medical devices used for treatment or monitoring.  \\
\midrule
User & What is shown in this image?\\
\llavaname{} & The image is a chest X-ray (CXR) that shows bilateral patchy infiltrates, which are areas of increased opacity in the lungs. These infiltrates can be indicative of various lung conditions, such as infections, inflammation, or other lung diseases.
\\
User & Are there any devices implanted in this patient? \\
\llavaname{} & Yes, it appears that the patient has several devices implanted. These devices may include endotracheal tubes, central venous catheters, and/or pulmonary artery catheters. These devices are typically used for monitoring and managing various medical conditions, as well as for providing support during surgical procedures.
\\
\midrule
User & What is shown in this image?  \\
GPT-4V & This chest X-ray image displays various medical devices, including screws and possibly a stabilizing plate near the clavicle, indicating past surgical interventions, as well as a central venous catheter used for medication or fluid administration. The image also includes a measurement (6.59 cm) highlighted in yellow, and the heart and lung fields are visible, showing patchy areas that might suggest medical conditions or normal variations. Red arrows likely point to areas of specific interest or concern, but accurate interpretation requires a professional medical evaluation.  \\
User &  Are there any devices implanted in this patient?  \\
GPT-4V  & Yes, the X-ray shows several implanted devices in the patient. There are screws and possibly a stabilizing plate near the clavicle area, which are likely related to previous surgical interventions. Additionally, there is a central venous catheter visible, which is used for administering medication or fluids directly into the bloodstream. These devices are common in patients undergoing complex medical treatments or surgeries.  \\
\midrule
\rowcolor{Gray}
User & What is shown in this image?\\
\rowcolor{Gray}
\shortname{} & The chest x-ray shows a patient with \textcolor{orange}{worsening ground-glass opacities in the lungs}. Ground-glass opacities are hazy areas seen on the x-ray that indicate a possible lung abnormality. \textcolor{orange}{In this case, the patient's condition seems to be deteriorating, as the opacities have worsened since admission}. The relative subpleural sparing suggests that \textcolor{orange}{the lung tissue near the outer surface of the lungs (close to the pleura) is less affected than other areas}. This finding may be relevant to the patient's underlying condition and the appropriate course of treatment. \textcolor{orange}{It is important to consult a healthcare professional for a thorough evaluation and proper diagnosis of the patient's condition}.
\\
\rowcolor{Gray}
User & Are there any devices implanted in this patient? \\
\rowcolor{Gray}
\shortname{}{} & Yes, the patient has an endotracheal tube implanted. This tube is inserted through the mouth or nose and into the trachea (windpipe) to help maintain an open airway and assist with breathing. The chest x-ray shows the position of the endotracheal tube, which is marked by a yellow line. \\
\bottomrule
\end{tabular}
}
\vspace{1mm}
\captionof{table}{Illustrative comparison of open generation quality on biomedical VQA. We also use the same prompt to directly interact with GPT-4V, using it as a performance standard. (\textcolor{orange}{orange} highlights \shortname{}'s capability in generating more detailed contextual information than the others)}
% briefly say what to notice.
% The prompt requires image understanding.}  
\vspace{-9mm}
\label{tab:visual_chat_example}  
  \end{minipage}
\end{table}

Table~\ref{tab:results_on_three_datasets} presents a comparative analysis of various methods on three key biomedical visual question answering (VQA) datasets: VQA-RAD, SLAKE, and VQA-PATH. The table specifically emphasizes zero-shot results to underline the generalization capabilities of each method. This focus provides valuable insights into how well these methods adapt to new, unseen data within the biomedical domain, a critical measure of their practical applicability and effectiveness.

\paragraph{Existing SoTA Methods.} 
% For the VQA-RAD dataset, the highest performance is observed in \citep{liu2023q2atransformer} for the Ref (79.19) and Closed (81.20) categories. In the SLAKE dataset, the best performance in the Ref and Closed categories is recorded by \citep{zhang2023large} with scores of 82.05 and 89.70, respectively. For the VQA-PATH dataset, \citep{bazi2023vision} and \citep{liu2023q2atransformer} demonstrate strong performance with scores of 85.61 and 88.85 in Closed questions. Nevertheless, traditional methods, such as those by~\citep{bazi2023vision} and~\citep{liu2023q2atransformer}, report high performance on specific tasks but often treat open-ended questions as classification problems. This approach can inflate generalizability as test answers frequently overlap with training data, which does not reflect true model robustness in real-world applications where questions and answers are more varied and less predictable.
For the VQA-RAD dataset, the highest performance metrics are reported in \citet{liu2023q2atransformer}, achieving scores of 79.19 in the Ref category and 81.20 in the Closed category. Similarly, in the SLAKE dataset, top results are recorded by \citet{zhang2023large} with scores of 82.05 and 89.70 in the Ref and Closed categories, respectively. In the VQA-PATH dataset, strong performance is demonstrated by \citet{bazi2023vision} and \citet{liu2023q2atransformer} in the Closed questions category, with scores of 85.61 and 88.85, respectively. However, traditional methods like those by VL Encoder-Decoder~\citep{bazi2023vision} and Q2ATransformer~\citep{liu2023q2atransformer} tend to treat open-ended questions as classification tasks. While this approach yields high performance on specific tasks, it can overstate the models’ generalizability. This is because test answers often overlap with training data, giving a skewed impression of model robustness. Such practices do not adequately reflect the true capabilities of these models in real-world scenarios, where questions and answers are more diverse and unpredictable.

\paragraph{Supervised Fine-tuning Results.}
When employing vanilla LLaVA methods without biomedical-specific adaptations, we observed improved performance after nine epochs of fine-tuning, particularly in closed-domain tasks across all datasets. Models trained with BioMed CLIP generally outperformed other approaches, highlighting the advantages of domain-specific training. Notably, compared to domain-agnostic LLaVA, the \llavaname{} equipped with the 7B (BioMed CLIP) model consistently demonstrated superior results. This model achieved high scores on VQA-RAD, SLAKE, and VQA-PATH datasets, underscoring its effectiveness in biomedical contexts pre-training and confirming its robustness in handling diverse medical data challenges.
% For vanilla LLaVA methods without specified biomedical adaption, after nine epochs of fine-tuning, model shows improved performance, particularly in closed-domain tasks across all datasets. The models trained with BioMed CLIP generally outperform others, suggesting the benefit of domain-specific training. Compared to LLaVA, \llavaname{} consistently demonstrates superior results, with the 7B (BioMed CLIP) model achieving high scores across VQA-RAD, SLAKE, and VQA-PATH, indicating its effectiveness in biomedical contexts.
% In the 1 epoch fine-tuning, the \llavaname{} 7B (CLIP) model reaches an average of 46.73, showing moderate improvement over zero-shot performance. With 9 epochs of fine-tuning, the \llavaname{} 7B (BioMed CLIP) model shows significant improvement, particularly in closed questions (average 75.40), indicating the effectiveness of extended training. The \llavaname{} 13B (CLIP) also demonstrates strong performance with an average of 74.05, suggesting that model size and fine-tuning duration are critical factors. Post fine-tuning, \llavaname{} 7B (BioMed CLIP) achieves the highest average (75.40) in the SLAKE dataset, underscoring the benefits of domain-specific pretraining and extensive fine-tuning. For the VQA-PATH dataset, \llavaname{} 13B (CLIP) achieves the highest closed score (92.39), indicating that larger models significantly benefit from fine-tuning.

\paragraph{Zero-shot Results.}
Our \shortname{} consistently achieves the highest zero-shot scores across all datasets and question types, demonstrating superior generalization capabilities—with an average improvement of over 10\%. However, an exception is noted in the open questions for VQA-RAD. The LLaVA family serves as the baseline, with performance averages ranging from 35.23 to 40.25. In comparison, models equipped with BioMedCLIP~\citep{zhang2023large} and Llama3-Med 8B generally excel in both open and closed questions, underscoring the value of domain-specific pre-training and larger foundational models. Notably, \llavaname{} 7B (BioMed CLIP) registers the highest zero-shot performance in VQA-RAD's open questions with a score of 37.84, although the closed questions score at 60.66 suggests some variability likely due to evaluation randomness. In the SLAKE dataset, both \llavaname{} 7B (w/ Mistral) and \shortname{} 8B show strong results, with the latter achieving the highest scores in open (44.34) and closed (61.06) questions. Impressively, our \shortname{} reaches an overall score of 46.22, nearly matching the 1-epoch supervised fine-tuning performance of \llavaname{} 7B using a domain-agnostic CLIP encoder.

\paragraph{Discussions.}
The \shortname{} 8B model's exceptional zero-shot performance significantly surpasses that of other models across all datasets and question types, with the exception of VQA-RAD, where performance plateaued. This plateau is conjectured to be due to the limited amount of training data, particularly the mere 22 Chest X-Ray images added by our proposed dataset. The model's success is attributed to its comprehensive training on a diverse and contextually rich biomedical dataset, which markedly enhances its ability to generalize effectively to new and unseen data. In contrast to traditional methods, which may overstate generalizability due to overlap between training and testing datasets, the \shortname{} 8B model demonstrates genuine robustness and adaptability, crucial for real-world applications. This superior generalization capability is particularly vital in dynamic and unpredictable environments like healthcare, where models must accurately interpret and respond to a broad spectrum of biomedical queries. The combination of advanced image encoding techniques, large-scale specialized datasets, and sophisticated model architecture not only positions \shortname{} 8B as a leading solution for biomedical visual question answering but also sets a new benchmark for future research and development in the field. Additionally, these results highlight the critical importance of sufficient and diverse training data in achieving comprehensive model performance across various tasks.

\begin{table}[t]
\begin{minipage}{0.99\textwidth}
\centering  
\small
\vspace{-5mm}
\scalebox{0.83}{
\begin{tabular}{p{2.1cm} p{13.2cm}  }
\toprule
\multicolumn{2}{l}{\bf CT example from VQA-RAD~\citet{Lau2018DescriptorA}}  \\
\midrule
% Caption & Chest x-ray (CXR) on Day 2 of admission post-intubation (yellow line showing the level of the endotracheal tube). Rapidly worsening ground-glass opacities are seen throughout the lungs with relative subpleural sparing (red arrows). \citep{bharat2020vaping}
% &  \includegraphics[height=3.5cm]{figures/example_imgs/visual_chat_example.jpg} \\
&  \includegraphics[height=4cm]{img-examples/ct-1.jpg} \\
User & What is shown in this image?  \\
GPT-4V & This image is a CT (Computed Tomography) scan of a human brain. CT scans use X-rays to create detailed cross-sectional images of the body's internal structures. This particular image shows various brain structures, such as the ventricles (the darker areas in the center), which are fluid-filled spaces in the brain, and the surrounding brain tissue. The scan provides a clear view of the brain's anatomy, and it can be used for diagnosing issues such as tumors, hemorrhages, or other abnormalities. \\ \midrule
\rowcolor{Gray}
 User & What is shown in this image?  \\
 \rowcolor{Gray}
\shortname{} &  This image is a computed tomography (CT) scan of the head, which shows a \textcolor{orange}{subdural hematoma in the left hemisphere}. A subdural hematoma is a type of bleeding that occurs between the layers of the brain's protective covering, specifically between the dura mater and the arachnoid mater. This can result from \textcolor{orange}{head injury, brain surgery, or other causes}. The image also shows a midline shift, which indicates that the hematoma is causing pressure on the brain and pushing it to the right side. Additionally, there is a small amount of left-sided subdural hygroma, which is a collection of fluid in the same area as the hematoma. \textcolor{orange}{The fluid collection is isodense to the brain, meaning it has a similar density to the brain tissue on the CT scan}. The image does not show any signs of cerebral edema, which is swelling of the brain. \textcolor{orange}{The total intracranial volume is normal, and there is no evidence of a mass lesion or an acute ischemic event}. The ventricular system appears to be normal, and there is no midline shift in the posterior fossa. The image does not show any acute changes, but it is important to monitor the patient's condition and consult with a healthcare professional for further evaluation and appropriate treatment.
 \\
\bottomrule
\end{tabular}
}
\vspace{1mm}
\captionof{table}{Illustrative comparison of open generation quality on CT image with GPT-4V.}
\vspace{-5mm}
\label{tab:visual_chat_example_ct}  
  \end{minipage}
\end{table}

\subsection{Quality of Generation}
Table~\ref{tab:visual_chat_example} showcases responses from various models (LLaVA~\citep{liu2023visual}, \llavaname{}~\citep{Li2023LLaVAMedTA}, GPT-4V~\citep{gpt4}, and \shortname{}) to biomedical visual questions based on a public chest X-ray image. Vanilla LLaVA's responses are generally vague, identifying the image merely as a medical illustration and noting the presence of medical devices without detailed specificity. \llavaname{} provides more precise medical descriptions, recognizing the image as a chest X-ray featuring bilateral patchy infiltrates, and notes the potential presence of endotracheal tubes and central venous catheters. GPT-4V offers a highly precise and context-rich response, detailing the intubation, ground-glass opacities, and an endotracheal tube. \shortname{}, however, excels in contextual and clinical relevance, describing the progression of ground-glass opacities, subpleural sparing, and emphasizing the need to consult a healthcare professional. It also accurately identifies the endotracheal tube and its placement. Overall, \shortname{} demonstrates \emph{superior response quality, providing comprehensive medical insights and context-specific information}, closely followed by GPT-4V. Additional comparative examples are detailed in Table~\ref{tab:visual_chat_example_path} (a pathology example) and~\ref{tab:visual_chat_example_ct} (a CT example).

\section{Related Work}
\label{sec:preliminary}

\paragraph{Biomedical Multi-Modal Models.}
The tremendous achievements of ChatGPT~\citep{chatgpt} and GPT-4~\citep{gpt4} have spurred the creation of several biomedical language models (LLMs) and chatbots, including ChatDoctor~\citep{yunxiang2023chatdoctor}, Med-Alpaca~\citep{han2023medalpaca}, PMC-LLaMA~\citep{wu2023pmc}, Clinical Camel~\citep{clinical_camel}, DoctorGLM~\citep{xiong2023doctorglm}, and Huatuo~\citep{wang2023huatuo}. Very recent work such as Visual Med-Alpaca~\citep{wu2023_visual-med_alpaca}, LLaVA~\citep{liu2023visual} and its specialized biomedical variant \llavaname{}~\citep{Li2023LLaVAMedTA} show promising results on building large vision-language foundation models (VLMs). All these models are built upon open-source LLMs and fine-tuned with specialized biomedical instruction-following datasets. This customization enables these models to effectively support various biomedical contexts by comprehending patient needs and providing knowledgeable advice.
Our proposed \shortname{} is inspired by both Visual Med-Alpaca~\citep{wu2023_visual-med_alpaca} and \llavaname{}~\citep{liu2023visual}. Although this two-stage training benefit the stable fine-tuning, the low resolution of the CLIP image encoder these models utilize can hindle model's understanding capability of granular biomedical image signals~\citep{McKinzie2024MM1MA}. As such, following the spirit of~\citet{Shi2024WhenDW}, instead of using a low resolution image encoder, we implement a hierarchical representation learning approach that optimizes the performance of existing pre-trained vision models. Instead of increasing the model size, we enhance their effectiveness by processing higher resolution images composed of multiple image scales. It is effective and efficient as existing image encoders are readily to use without further finetuning on high resolution images.

\paragraph{Biomedical Visual Question Answering (VQA).}
Biomedical visual question understanding is of great value in clinics. Existing biomedical VQA methods fall into two main categories: discriminative and generative approaches.
Discriminative methods approach VQA as a classification task, where models predict answers from a predefined set. While these methods perform well, they are confined to closed-set predictions~\citep{he2020pathvqa} and require adjustments when a custom answer set is needed during inference~\citep{li2022self,zhang2023large,eslami2023pubmedclip}. This approach is not ideal for creating a versatile biomedical assistant capable of addressing open-ended questions in real-world scenarios.
To overcome this limitation, generative methods have been developed to produce answers as free-form text sequences~\citep{bazi2023vision,liu2023q2atransformer,van2023open}. These methods are more flexible, as they can naturally handle closed-set questions when candidate answers are provided in the language instructions. Moreover, \llavaname{}~\citep{Li2023LLaVAMedTA} utilizes GPT-4 to self-instruct biomedical multi-modal instruction-following data using freely available, broad-coverage biomedical image-text pairs extracted from PubMed Central~\citep{zhang2023large}. In this work, we follow the similar procedure to generate more diverse instruction datasets but using LLaMA3 70B~\citep{touvron2023llama}.

\paragraph{Architecture Design.}
We employ \shortname{} with the model architecture akin to the prefix tuning of language models (LMs) proposed by~\citep{van2023open,liu2023visual}, wherein a new trainable module bridges a frozen image encoder with a causal LM. In~\citep{van2023open}, a three-layer MLP network is utilized to transform visual features into a visual prefix, leveraging pre-trained LMs such as GPT2-XL~\citep{radford2019language}, BioMedLM~\citep{venigalla2022biomedlm}, and BioGPT~\citep{luo2022biogpt}, with sizes ranging from 1.5B to 2.7B parameters. In contrast, \shortname{} employs a linear projection and a 7B LM~\citep{vicuna,touvron2023llama}. Similar to these models, ~\citet{Li2023LLaVAMedTA} design connectors following the same spirit. Interestingly, thought there are many other different connector types, researchers from Apple~\citep{McKinzie2024MM1MA} observe the lesson from large scale ablation studies that \textit{number of visual tokens and image resolution matters most, while the type of VL connector has little effect}. Consequently, in this work, we focus on the fine-grained biomedical image signal understanding by adopting a standard MLP as the connector.

\vspace{-2mm}
\section{Conclusion}
\label{sec:conclusion}

This paper has investigated the integration of vision-language modeling in the biomedical field, introducing a novel instruct dataset enriched with medical image-text pairs generated by the Claude3-Opus and LLaMA3 70B models, an innovative image encoding strategy with hierarchical representations, and the Llama3-Med model, which demonstrates state-of-the-art performance on biomedical VQA benchmarks. These advancements underscore the capability of multimodal models to serve as robust, versatile assistants in healthcare, effectively overcoming the challenges of biomedical image-text integration. The enhanced dataset and encoding techniques improve the model’s ability to decipher complex biomedical imagery, providing tools that are more accurate and contextually relevant for medical professionals. Looking ahead, future research will aim to refine model performance through continuous instruction-tuning, domain-specific pretraining, and the expansion of biomedical datasets, thereby enhancing the impact of AI in healthcare.

\bibliography{neurips_2024.bib}
% \bibliography{egbib}
% \bibliographystyle{plain}
\bibliographystyle{abbrvnat}

%%%%%%%%%%%%%%%%%%%%%%%%%%%%%%%%%%%%%%%%%%%%%%%%%%%%%%%%%%%%
% \newpage
\appendix
\newpage
\section{Data}
\subsection{Datasets used in \shortname{}}

\paragraph{Biomedical Images.} All the images are collected from PMC-15M~\citep{zhang2023large} with downloading urls as following (structured by \llavaname{}~\citep{Li2023LLaVAMedTA}):

\url{https://hanoverprod.z21.web.core.windows.net/med_llava/llava_med_image_urls.jsonl}

\paragraph{Alignment Pre-training.} In this work, we adopt \llavaname{}'s \texttt{alignment\_500k} for the connector alignment pretraining as we focus on synthesizing more data for conversational instruction fine-tuning:

\url{https://hanoverprod.z21.web.core.windows.net/med_llava/alignment/llava_med_alignment_500k.json}

\paragraph{Supervised Fine-tuning.} Following the CC BY NC 4.0 license and a similar data synthesis procedure as~\llavaname{}, we produce \texttt{llama3\_med\_ft\_20k}:

\url{https://github.com/standardmodelbio/Llama3-Med/releases/download/v0.1/llama3_med_instruct_finetuning_llama3_claude3_20k.json}

The finalized dataset used for training \shortname{} is a combination of \texttt{LLaVA-Med 60K-IM} and \texttt{\shortname{} FT-20K}, of which \texttt{LLaVA-Med 60K-IM} can be downloaded here:

\url{https://hanoverprod.z21.web.core.windows.net/med_llava/instruct/llava_med_instruct_60k_inline_mention.json}

\subsection{Prompts for LLaMA3 70B and Claude3-Opus to Generate Data}
Compared to~\llavaname{}, we further optimize the generation instruction to be more concise, maintaining clarity, and ensure the LLMs adhere to the specified requirements while generating the conversation, as Table~\ref{tab:prompts} illustrates. By giving a clear objective or task to the LLM, the generation could be improved by empirical studies. Table~\ref{tab:generated_example} shows an example of the LLM's generation.

\begin{table}[t]
\begin{minipage}{0.99\textwidth}
\centering  
\small
\vspace{-5mm}
\scalebox{0.83}{
\begin{tabular}{p{2.1cm} p{13.2cm}  }
\toprule
\multicolumn{2}{l}{\bf Prompting LLaMA3 and Claude3-Opus to Synthesize Data }  \\
\midrule
\multirow{2}{*}{\begin{tabular}[c]{@{}l@{}}Prompt used \\ by~\llavaname{}\end{tabular}} & You are an AI assistant specialized in biomedical topics.

You are provided with a text description (Figure Caption) of a figure image from a biomedical
research paper. In some cases, you may have additional text (Figure Context) that mentions
the image. Unfortunately, you don’t have access to the actual image.

Below are requirements for generating the questions and answers in the conversation:

- Avoid quoting or referring to specific facts, terms, abbreviations, dates, numbers, or
names, as these may reveal the conversation is based on the text information, rather than
the image itself. Focus on the visual aspects of the image that can be inferred without
the text information.

- Do not use phrases like "mentioned", "caption", "context" in the conversation. Instead,
refer to the information as being "in the image."

- Ensure that questions are diverse and cover a range of visual aspects of the image.

- The conversation should include at least 2-3 turns of questions and answers about the
visual aspects of the image.

- Answer responsibly, avoiding overconfidence, and do not provide medical advice or
diagnostic information. Encourage the user to consult a healthcare professional for
advice. \\ \midrule
\rowcolor{Gray}
 Our Prompt & 
You are an AI assistant specialized in biomedical topics.

You are provided with a text description (Figure Caption) of a figure image from a biomedical research paper. In some cases, you may have additional text (Figure Context) that mentions the image. Unfortunately, you don’t have access to the actual image.

\textcolor{orange}{Your task is to generate questions and answers about the visual aspects of the image based on the provided description, which adhering to the following guidelines:}

- \textcolor{violet}{Focus on the visual aspects of the image that can be inferred without referring to specific facts, terms, abbreviations, dates, numbers, or names.}

- Avoid using phrases like "mentioned", "caption", or "context". Refer to the information as being "in the image".

- Ensure questions are diverse and cover a range of visual aspects of the image.

- Include at least 2-3 turns of questions and answers about the visual aspects of the image.

- Answer responsibly, avoiding overconfidence, and do not provide medical advice or diagnostic information. Encourage the user to consult a healthcare professional for advice.

\textcolor{orange}{Now, generate sample conversations based on these guidelines.}
  \\
\bottomrule
\end{tabular}
}
\vspace{1mm}
\captionof{table}{Prompts used for both LLaMA3 and Claude3 to generate instruction fine-tuning data. Following~\llavaname{}, we also include curated few-shot examples for better generation quality, where each example consists of an input sample (\emph{context}) and an output sample (\emph{response}). \textcolor{orange}{orange} represents the added difference. \textcolor{violet}{violet} represents the optimized difference.}
% \vspace{-5mm}
\label{tab:prompts}  
  \end{minipage}
\end{table}

\begin{table}[t]
\begin{minipage}{\textwidth}
\centering  
\small
\vspace{-5mm}
\scalebox{0.85}{
\begin{tabular}{p{2.1cm} p{13.2cm}  }
\toprule
\multicolumn{2}{l}{\bf Sample Input and Output using Prompts above }  \\
\midrule
\multirow{2}{*}{Sample Input} & \textbf{Figure Caption}: Magnetic resonance imaging (MRI) prior to chemotherapy. MRI revealed extraskeletal mass of distal femur (black arrow), and a lesion thought to be hemorrhage and dissemination of tumor tissues (arrow) were observed \\
 & \textbf{In Context Mentioning \#1}: Magnetic resonance imaging ( MRI ) revealed iso-signal intensity on T1-weighted images and high-signal intensity on T2-weighted images of the left distal femur ; it also revealed an extraskeletal mass ( Fig. 2 ) . \\
 & \textbf{In Context Mentioning \#2}: Furthermore , MRI revealed diffuse signal alteration in the muscles and the neurovascular areas surrounding the lesion in the distal femur ; hence , hemorrhage and dissemination of the tumor were considered ( Fig. 2 ) .
\\ \midrule
\rowcolor{Gray}
 Sample Output & \textbf{User}: \texttt{<image>}{\textbackslash n}What is the location of the extraskeletal mass? \\
 \rowcolor{Gray}
 & \textbf{Assistant}: The extraskeletal mass is located at the distal femur, which is the lower end of the thigh bone near the knee joint. \\
 \rowcolor{Gray}
 & \textbf{User}: What are the arrows pointing to? \\
 \rowcolor{Gray}
 & \textbf{Assistant}: The black arrow is pointing to the extraskeletal mass of the distal femur, while the other arrow is pointing to a lesion that is thought to be hemorrhage and dissemination of tumor tissues. \\
 \rowcolor{Gray}
 & \textbf{User}: What can you say about the signal intensities on T1-weighted and T2-weighted images? \\
 \rowcolor{Gray}
 & \textbf{Assistant}: The MRI revealed iso-signal intensity on T1-weighted images and high-signal intensity on T2-weighted images of the left distal femur. Iso-signal intensity means that the signal intensity of the area of interest is similar to that of the surrounding tissue, while high-signal intensity indicates that the area of interest appears brighter compared to the surrounding tissue. These signal intensities can provide information about the nature of the tissue and help in identifying abnormalities, such as tumors or inflammation.
  \\
\bottomrule
\end{tabular}
}
\vspace{1mm}
\captionof{table}{One of the examples of synthesized \texttt{\shortname{} FT-20K}. The "In Context Mentioning" in the input offers additional information by including sentences from the corresponding articles that reference the figure.}
% \vspace{-5mm}
\label{tab:generated_example}  
  \end{minipage}
\end{table}

\section{Limitations}
\label{sec:limit}
While our work demonstrates significant advancements in the development of multimodal biomedical models, there are several limitations that need to be addressed. Firstly, despite the comprehensive dataset we curated, the performance of our model on the VQA-RAD dataset did not show significant improvement, highlighting the limitations of the training data for this specific task. This indicates the necessity for even larger and more diverse datasets to cover the full spectrum of biomedical imagery and associated textual data. Secondly, although our model leverages sophisticated image encoding techniques and advanced language models, it still relies heavily on the quality and variety of the input data. Any biases or gaps in the dataset can directly affect the model's performance and generalization capabilities. Lastly, the computational resources required for training large-scale models like LLaMA3-Med are substantial, which may not be feasible for all research institutions, potentially limiting the accessibility and reproducibility of our methods. Future work should focus on optimizing these models to reduce computational overhead while maintaining performance, as well as exploring methods to mitigate biases inherent in the training data.

\section{Broader Impact}
\label{sec:impact}
The advancements presented in this work have the potential to significantly impact the field of biomedical research and healthcare. By developing models capable of accurately interpreting and responding to complex biomedical queries, we can enhance the capabilities of clinical decision support systems, leading to more informed and timely decisions in patient care. This can improve diagnostic accuracy and treatment planning, ultimately leading to better patient outcomes. Moreover, the ability to perform zero-shot generalization across various biomedical tasks means that these models can be applied in diverse and dynamic clinical settings without extensive retraining, making them highly adaptable tools for medical professionals.
Additionally, the methodologies and datasets we developed can serve as valuable resources for the broader AI and biomedical communities, fostering further research and innovation in multimodal learning. By addressing the limitations and biases in current models, our work also contributes to the ongoing efforts to create more equitable and reliable AI systems in healthcare. However, it is crucial to consider the ethical implications and ensure that these technologies are used responsibly, with appropriate oversight to protect patient privacy and data security. Overall, the broader impact of our work lies in its potential to enhance the integration of AI in medicine, driving forward the frontiers of medical research and clinical practice.

\section{Implementation Details for Llama3-Med}
\label{sec:appendix}

Llama3-Med is implemented using a two-stage training process adapted from LLaVA\footnote{https://llava-vl.github.io/}. The initial stage, \textit{feature alignment}, utilizes a subset of 600K images from the PMC-15M dataset~\footnote{https://github.com/microsoft/LLaVA-Med} to integrate a frozen pretrained vision encoder with a frozen large language model (LLM). The subsequent stage, \textit{visual instruction tuning}, involves training the model to follow multimodal instructions using 60K (from \llavaname{}) GPT-generated multimodal instruction-following samples, supplemented by approximately 20K instruction biomedical QA pairs we produced using Claude3 and LLaMA3 70B also from PMC-15M.

The model is trained on 4 NVIDIA H100 GPUs, each with 80GB of memory. With fewer GPUs, setups can be adjusted accordingly include reducing the \texttt{per\_device\_train\_batch\_size} and increasing the \texttt{gradient\_accumulation\_steps} to maintain the global batch size as: \texttt{per\_device\_train\_batch\_size} $\times$ \texttt{gradient\_accumulation\_steps} $\times$ \texttt{num\_gpus}.

\subsection{Models}
\paragraph{CLIP Image Encoder.} We use Apple's pre-trained CLIP image encoder\footnote{https://huggingface.co/apple/DFN5B-CLIP-ViT-H-14-378}, which is a CLIP (Contrastive Language-Image Pre-training) model trained on DFN-5B. Data Filtering Networks (DFNs) are small networks used to automatically filter large pools of uncurated data. This model was trained on 5B images that were filtered from a pool of 43B uncurated image-text pairs (12.8B image-text pairs from CommonPool-12.8B + 30B additional public image-text pairs). It supports image resolution natively for 378$\times$ 378.

\paragraph{Mistral.} We use Mistral-7B-Instruct-v0.2\footnote{https://huggingface.co/mistralai/Mistral-7B-Instruct-v0.2} as the LLM backbone, which is an instruct fine-tuned version of the Mistral-7B~\citep{Jiang2023Mistral7}.

\paragraph{LLaMA3.} Name of the game, we leverage the SoTA open source LLM Meta-Llama-3-8B-Instruct\footnote{https://huggingface.co/meta-llama/Meta-Llama-3-8B-Instruct}. Meta developed and released the Meta Llama 3 family of LLMs, a collection of pretrained and instruction tuned generative text models in 8 and 70B sizes. In this work, we focus on 8B version given the infrastructure limitation and training efficiency. 
% The Llama 3 instruction tuned models are optimized for dialogue use cases and outperform many of the available open source chat models on common industry benchmarks. Further, in developing these models, we took great care to optimize helpfulness and safety.
Specifically, the final image resolution we utilize is 1,134 which is three times more of 378. Basically we concatenate the image embeddings of resolution 378, 756 and 1,134. Resolution higher than this would intrinsically increase the memory capacity requirements and further training costs.

\subsection{Hyperparameters}
\begin{table}[h]
\centering
\begin{tabular}{|l|l|}
\hline
\textbf{Hyperparameter} & \textbf{Value} \\ \hline
Global Batch Size & 256 \\ \hline
Learning Rate & 1e-3 \\ \hline
Epochs & 1 \\ \hline
Max Length & 2048 \\ \hline
Weight Decay & 0 \\ \hline
Warmup Ratio & 0.03 \\ \hline
\end{tabular}
\vspace{1mm}
\caption{Pretraining hyperparameters for \shortname{}}
\label{tab:pretraining-hyperparameters}
\end{table}

\begin{table}[h]
\centering
\begin{tabular}{|l|l|}
\hline
\textbf{Hyperparameter} & \textbf{Value} \\ \hline
Global Batch Size & 128 \\ \hline
Learning Rate & 2e-5 \\ \hline
Epochs & 1 \\ \hline
Max Length & 2048 \\ \hline
Weight Decay & 0 \\ \hline
Warmup Ratio & 0.03 \\ \hline
\end{tabular}
\vspace{1mm}
\caption{Finetuning hyperparameters for \shortname{}}
\label{tab:finetuning-hyperparameters}
\end{table}

\subsection{Training Logs}
The entire training logs can be found at \url{https://api.wandb.ai/links/zachstark/y47cx29l}.

% \subsection{An example from pathology}

\begin{table}[t]
  \begin{minipage}{0.99\textwidth}
\centering  
\small
\vspace{-4mm}
\scalebox{0.83}{
\begin{tabular}{p{2.1cm} p{13.2cm}  }
\toprule
\multicolumn{2}{l}{\bf Pathological example from VQA-PATH~\citet{He2020PathVQA3Q}}  \\
\midrule
% Caption & Chest x-ray (CXR) on Day 2 of admission post-intubation (yellow line showing the level of the endotracheal tube). Rapidly worsening ground-glass opacities are seen throughout the lungs with relative subpleural sparing (red arrows). \citep{bharat2020vaping}
% &  \includegraphics[height=3.5cm]{figures/example_imgs/visual_chat_example.jpg} \\
&  \includegraphics[height=4cm]{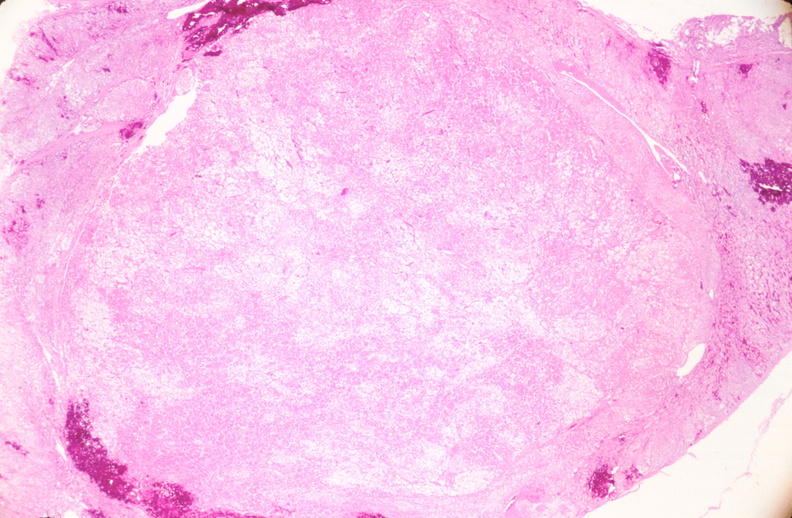} \\
User & What is shown in this image?  \\
GPT-4V & This image is a microscopic view of a tissue sample, specifically a histopathology slide stained with hematoxylin and eosin (H\&E). The slide shows a large, predominantly pink (eosinophilic) mass with areas of purple (hematoxylin staining) indicating cell nuclei. This type of slide is typically used to examine the microscopic structure of tissues to diagnose diseases, with the pattern, coloration, and structure providing clues about the type of tissue and any pathological changes. The specifics of the pathology, such as the type of tissue and the nature of any disease present, would require interpretation by a medical professional, especially a pathologist. \\ \midrule
\rowcolor{Gray}
 User & What is shown in this image?  \\
 \rowcolor{Gray}
\shortname{} &  The image is a histopathology slide, which is a microscopic examination of a tissue sample. In this case, the slide shows a necrotic tumor, which means that there is dead or dying tissue within the tumor. This can be seen as areas of \textcolor{orange}{tissue damage or discoloration in the image}. Additionally, the slide shows \textcolor{orange}{viable tissue at the periphery}, which indicates that some living cells are still present at the outer edges of the tumor. The presence of both necrotic and viable tissue can provide valuable information about the tumor's characteristics and behavior. Histopathology slides are often used by pathologists to study the nature of a tumor and to help determine the appropriate course of treatment. \\
\bottomrule
\end{tabular}
}
\vspace{1mm}
\captionof{table}{Illustrative comparison of open generation quality on pathology example with GPT-4V.}
\vspace{-5mm}
\label{tab:visual_chat_example_path}  
  \end{minipage}
\end{table}

\end{document}